\RequirePackage{fix-cm}
\documentclass[smallextended]{svjour3}       
\smartqed  
\usepackage{graphicx}
\usepackage{amsmath}
\usepackage{amsfonts}
\usepackage{longtable}
\usepackage{multirow}
\usepackage[colorlinks,linkcolor=blue,anchorcolor=blue,citecolor=blue]{hyperref}

\usepackage{color}

\usepackage{CJKutf8,xr-hyper}
\usepackage[breakable,fitting]{tcolorbox}
\newtcolorbox{mybox}{breakable,coltitle=black,colbacktitle=gray!20,colframe=black,colback=white,coltext=black}

\newcommand{\orcid}[1]{\href{https://orcid.org/#1}{\includegraphics[scale=1]{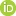}}}

\newcommand{\R}{\mathbb{R}}
\newcommand{\N}{\mathbb{N}}

\usepackage[ruled,linesnumbered]{algorithm2e}
\newcommand{\argmin}{\text{argmin}}
\newcommand{\argmax}{\text{argmax}}
\newcommand{\IntEnt}[1]{\left[\!\left[#1\right]\!\right]}
\DeclareMathOperator*{\dom}{dom}

\journalname{Optimization Letters}
\begin{document}

\title{A Difference-of-Convex Programming Approach With Parallel Branch-and-Bound For Sentence Compression Via A Hybrid Extractive Model\thanks{\textbf{Funding:} the authors are supported by the National Natural Science Foundation of China (Grant 11601327) and the National ``$985$" Key Program of China (Grant WF220426001).}
}

\titlerunning{PDCABB for sentence compression}        

\author{Yi-Shuai Niu \orcid{0000-0002-9993-3681} \and Yu You \and Wenxu Xu \and Wentao Ding \and Junpeng Hu \and Songquan Yao}

\authorrunning{Y.S. Niu et al.} 

\institute{Yi-Shuai Niu \at
	School of Mathematical Sciences \& SJTU-Paristech Elite Institute of Technology, Shanghai Jiao Tong University, China.
	\email{niuyishuai@sjtu.edu.cn}           
	\and
	Yu You, Wenxu Xu, Wentao Ding, Junpeng Hu \at School of Mathematical Sciences, Shanghai Jiao Tong University, China. 
	\email{\{youyu0828,xuwx999,569018166,hjp3268\}@sjtu.edu.cn}	
	\and
	Songquan Yao \at School of Mechanical and Automotive Engineering, Shanghai University of Engineering Science, China.
	\email{songquanyao@163.com}
}

\date{Received: date / Accepted: date}

\maketitle

\begin{abstract}
Sentence compression is an important problem in natural language processing with wide applications in text summarization, search engine and human-AI interaction system etc. In this paper, we design a hybrid extractive sentence compression model combining a probability language model and a parse tree language model for compressing sentences by guaranteeing the syntax correctness of the compression results. Our compression model is formulated as an integer linear programming problem, which can be rewritten as a Difference-of-Convex (DC) programming problem based on the exact penalty technique. We use a well-known efficient DC algorithm -- DCA to handle the penalized problem for local optimal solutions. Then a hybrid global optimization algorithm combining DCA with a parallel branch-and-bound framework, namely PDCABB, is used for finding global optimal solutions. Numerical results demonstrate that our sentence compression model can provide excellent compression results evaluated by F-score, and indicate that PDCABB is a promising algorithm for solving our sentence compression model. 

\end{abstract}

\section{Introduction}\label{sec:introduction}
In recent years, due to the rapid development of artificial intelligence (AI) technology, and the need to process large amounts of natural language information in very short response time, the problem of sentence compression has attracted researchers' attention. Tackling this problem is an important step towards natural language understanding in AI. Nowadays, there are various technologies involving sentence compression such as text summarization, search engine, question answering, and human-AI interaction systems. The general idea of sentence compression is to make a shorter sentence or generate a summarization for the original text by containing the most important information and maintaining grammatical rules.

Overall speaking, there are two categories of models for sentence compression: \emph{extractive models} and \emph{abstractive models}. Extractive models reduce sentences by extracting important words from the original text and putting them together to form a shorter one. Abstractive models generate sentences from scratch without being constrained to reuse words from the original text. 
	
On extractive models, the paper of Jing (\cite{paper_jing_2000} in 2000) could be one of the first works addressed on this topic with many rewriting operations as deletion, reordering, substitution, and insertion. This approach is realized based on multiple knowledge resources (such as WordNet and parallel corpora) to find the pats that can not be removed if they are detected to be grammatically necessary by using some simple rules. Later, Knight and Marcu (\cite{paper_knight_2002} in 2002) investigated discriminative models by proposing a decision-tree to find the intended words through a tree rewriting process. Hori and Furui (\cite{paper_hori_2004} in 2004) proposed a model for automatically transcribed spoken text using a dynamic programming algorithm. McDonald (\cite{paper_mcdonald_2006} in 2006) presented a sentence compression model using a discriminative large margin algorithm. He ranked each candidate compression using a scoring function based on the Ziff-Davis corpus and a Viterbi-like algorithm. The model has a rich feature set defined over compression bigrams including parts of speech, parse trees, and dependency information, without using a synchronous grammar. Clarke and Lapata (\cite{paper_clarke_2008} in 2008) reformulated McDonald's model in the context of integer linear programming (ILP) and extended with constraints to ensure (based on probability) that the compression results are grammatically and semantically well formed. The corresponding ILP model is solving in using the branch-and-bound algorithm. Recently, Google guys use LSTM recurrent neural networks (RNN)  (\cite{paper_filippova_2015} in 2015) to generate shorter sentences. 
		 
On abstractive models, a noisy-channel machine translation model was proposed by Banko et al. (\cite{paper_banko_2000} in 2000), then Knight and Marcu (\cite{paper_knight_2002}) indirectly using the noisy-channel model to  construct a compressed sentence from some scrambled words based on the probability of mistakes. Later the noisy-channel based model is formalized on the task of abstractive sentence summarization around the DUC-2003 and DUC-2004 competitions by Zajic et al. (\cite{paper_zajic_2004} in 2004) and Over et al. (\cite{paper_over_2007} in 2007). Later, Cohn and Lapata (\cite{paper_cohn_2008} in 2008) proposed systems which made heavy use of the syntactic features of the sentence-summary pairs. Recently, Facebook guys (\cite{paper_rush_2015,paper_chopra_2016} in 2015) proposed attention-based neural network models for this problem using and showing the state-of-the-art performance on the DUC tasks.
	
In our paper, we propose a hybrid extractive model to delete words from the original sentences by guaranteeing the grammatical rules and preserving the main meanings. \textit{Our contributions} are focused on: (1) Establish a hybrid model (combining a \emph{Parse tree model} and a \emph{Probabilistic model}) in which the parse tree model can help to guarantee the grammatical correctness of the compression result, and the probabilistic model is used to formulate our task as an integer programming problem (ILP) whose optimal solution is related to a compression with maximum probability to be a correct sentence. To hybrid them, we use the parse tree model to extract the sentence truck, then fix the corresponding integer variables in the probabilistic model to derive a simplified ILP problem which can provide improved compression results comparing to the parse tree model and the probabilistic model. (2) Formulate the ILP model as a DC programming problem, and apply a mixed-integer programming solver PDCABB (an implementation of a hybrid algorithm combing DC programming approach -- DCA with a parallel branch-and-bound framework) developed by Niu (\cite{PDCABB,paper_niu_2018} in 2017) for solving our sentence compression model. As a result, our sentence compression model with PDCABB can often provide high quality compression result within a very short compression time.

The paper is organized as follows: Section \ref{sec:sentence_compression_models} is dedicated to establish hybrid sentence compression model. In Section \ref{sec:solvethemodel}, we will present DC programming approach for solving ILP. The numerical simulations with experimental setups will be reported in Section \ref{sec:experimental_result}. Some conclusions and future works will be discussed in the last section.

\section{Hybrid Sentence Compression Model}\label{sec:sentence_compression_models}
Our hybrid sentence compression model is based on an Integer Linear Programming (ILP) probabilistic model firstly proposed by Clarke and Lapata in \cite{paper_clarke_2008}, namely \emph{Clarke-Lapata ILP model}. We will combine it with a parsing tree model and take different sentence types into consideration in order to improve the quality and effectiveness of compression. In this section, we will firstly present the Clarke-Lapata ILP model and the parse tree model, and introduce our hybrid model at last.

\subsection{Clarke-Lapata ILP Model}\label{subsec:ILPmodel}

	Let us use a sequence $\vec{x} := (x_0, x_1, x_2, \ldots, x_n, x_{n+1})$ to  present a source sentence where $n \in \N^*$ is the number of the words, the leading element $x_0$ is the start token denoted by `start', the ending element $x_{n+1}$ is the end token denoted by `end', and the subsequence $(x_1,\ldots,x_n)$ is the list of words in the source sentence. The probability of $x_N$ followed by a sequence of words $x_{1},\ldots,x_{N-1}$ is denoted by $P(x_N~ |~ x_1,\ldots, x_{N-1})$, which can be computed by the classical $N$-gram language model based some text corpora. For English language, we often use unigram, bigram and trigram (i.e., $N=1,2,3$) since most of English phrases consist of less than $3$ words. For computing these probabilities in practice, there are many existing packages, e.g., the NLTK \cite{NLTK} package in Python. Clearly, the probability $P(x_N~ |~ x_1,\ldots, x_{N-1})$ is closely related to the context of the sentence. E.g.,  mathematical terms such as ``homomorphic" will appear more frequently (thus with higher probability) in articles of mathematics than sports. NLTK provides a convenient trainer to learn $N$-gram probability based on personal corpora. For missing words in corpora, smoothing techniques such as Kneser–Ney smoothing and Laplace smoothing are used to avoid zero probabilities of these missing words. The reader can refer to \cite{book_steven_2009} for more details about NLTK and $N$-gram model.
	
	Based on $N$-gram model, Clarke and Lapata proposed in \cite{paper_clarke_2008} a probabilistic sentence compression model which aims at extracting a subsequence of words in a sentence $\vec{x}$ with maximum probability to form a well-structured shorter sentence. Moreover, it is also suggested to introduce some constraints for restricting the set of allowable word combinations. Therefore, the sentence compression task is formulated as an optimization problem which consists of three parts: Decision variables, Objective function and Constraints.

\subsubsection{Decision Variables}
We associate to each word $x_i, i\in \IntEnt{1,n}$\footnote{$\IntEnt{m,n}$ with $m\leq n$ stands for the set of integers between $m$ and $n$.}
a binary variable $\delta_i$, called \emph{choice decision variable}, with $\delta_i=1$ if $x_i$ is in a compression and $0$ otherwise. In order to take context information into consideration, we also need to introduce the \emph{context decision variables} $(\alpha,\beta,\gamma)$ as:
$$\begin{array}{ll}
\alpha_i=
\begin{cases}
1& \text{if $x_i$ starts a compression}\\
0& \text{otherwise}
\end{cases}
&, \forall i \in \IntEnt{1,n}\par\vspace{3ex}\\
\beta_{ij}=
\begin{cases}
1& \text{if sequence $x_i$, $x_j$ ends a compression}\\
0& \text{otherwise}
\end{cases}
&, \begin{array}{l}
\forall i \in \IntEnt{1,n-1}\\
\forall j \in \IntEnt{i+1,n}
\end{array}\par\vspace{3ex}\\
\gamma_{ijk}=
\begin{cases}
1& \text{if sequence $x_i$, $x_j$, $x_k$ is in a compression}\\
0& \text{otherwise}
\end{cases}
&, \begin{array}{l}
\forall i \in \IntEnt{1,n-2}\\
\forall j \in \IntEnt{i+1,n-1}\\
\forall k \in \IntEnt{j+1,n}
\end{array}
\end{array}$$
There are totally $\frac{n^3+3n^2+14n}{6}$ binary variables in $(\delta,\alpha,\beta,\gamma)$.

\subsubsection{Objective Function} Our objective is to maximize the probability of the compression as:
\begin{eqnarray}
f(\alpha,\beta,\gamma) &:=& \sum_{i=1}^n\alpha_i P\left(x_i|\text{`start'}\right)+
\sum_{i=1}^{n-2}\sum_{j=i+1}^{n-1}\sum_{k=j+1}^n\gamma_{ijk} P\left(x_k|x_i, x_j\right) \nonumber\\
&& + \sum_{i=0}^{n-1}\sum_{j=i+1}^n\beta_{ij} P\left(\text{`end'}|x_i, x_j\right)\nonumber
\end{eqnarray}
where $P\left(x_i|\text{`start'}\right)$ stands for the probability of a sentence starting with $x_i$, $P\left(x_k|x_i, x_j\right)$ denotes the probability that $x_i,x_j,x_k$ successively occurs in a sentence, and $P\left(\text{`end'}|x_i, x_j\right)$ means the probability that $x_i,x_j$ ends a sentence. The probability $P\left(x_i|\text{`start'}\right)$ is computed by bigram model, and the others are computed by trigram model based on some corpora. The function $f$ indicates the probability of a compression associated with decision variables $(\alpha,\beta,\gamma)$. Among all possible compressions, we will find one with maximal probability.

\subsubsection{Constraints}
To ensure that the extracted sequence forms a sentence, we must introduce some constraints to restrict the possible combinations in $\vec{x}$.

\noindent{\bf Constraint 1} Exactly one word can begin a sentence.
\begin{equation}\label{cons:1}
\sum_{i=1}^n\alpha_i = 1.
\end{equation}

\noindent{\bf Constraint 2} If a word is included in a compression, it must either start the sentence, or be preceded by two other words, or be preceded by the `start' token and one other word.
\begin{equation}\label{cons:2}
\delta_k-\alpha_k-\sum_{i=0}^{k-2}\sum_{j=1}^{k-1}\gamma_{ijk} = 0, \forall k \in\IntEnt{1,n}.
\end{equation}

\noindent{\bf Constraint 3} If a word is included in a compression, it must either be preceded by one word and followed by another, or be preceded by one word and the `end' token.
\begin{equation}\label{cons:3}
\delta_j-\sum_{i=0}^{j-1}\sum_{k=j+1}^{n}\gamma_{ijk}-\sum_{i=0}^{j-1}\beta_{ij} = 0, \forall j \in \IntEnt{1,n}.
\end{equation}

\noindent{\bf Constraint 4} If a word is in a compression, it must either be followed by two words, or be followed by one word and the `end' token.
\begin{equation}\label{cons:4}
\delta_i-\sum_{j=i+1}^{n-1}\sum_{k=j+1}^{n}\gamma_{ijk}-\sum_{j=i+1}^{n}\beta_{ij}
-\sum_{h=0}^{i-1}\beta_{hi} = 0, \forall i \in \IntEnt{1,n}.
\end{equation}

\noindent{\bf Constraint 5} Exactly one word pair can end the sentence.
\begin{equation}\label{cons:5}
\sum_{i=1}^{n-1}\sum_{j=i+1}^n\beta_{ij} = 1.
\end{equation}

\noindent{\bf Constraint 6} The length of a compression should be bounded.
\begin{equation}\label{cons:6}
\underline{l}\leq  \sum_{i=1}^n\delta_i \leq \bar{l}.
\end{equation}

\noindent{\bf Constraint 7} The introducing term for preposition phrase (PP) or subordinate clause (SBAR) must be included in the compression if any word of the phrase is included. Otherwise, the phrase should be entirely removed. Let us denote $I_i = \{j: x_j\in \text{PP/SBAR}, j\neq i\}$ the index set of the words included in PP/SBAR leading by the introducing term $x_i$, then
	\begin{equation}\label{cons:7}
	\sum_{j\in I_i}\delta_j \geq \delta_i, \delta_i \geq \delta_j, \forall j\in I_i.
	\end{equation}
All supported abbreviations such as (PP, SBAR) are listed in Appendix \ref{appendix:A}.

\subsubsection{ILP Probabilistic Model}
As a conclusion, the Clarke's ILP probabilistic model for sentence compression is formulated as a binary linear program as:
	\begin{equation}\label{prob:ILP}
	\max\{f(\alpha,\beta,\gamma) ~|~ (\ref{cons:1})-(\ref{cons:7}), (\alpha,\beta,\gamma,\delta)\in\{0,1\}^{\frac{n^3+3n^2+14n}{6}} \}.
	\end{equation}
	with $O(n^3)$ binary variables and $O(n)$ linear constraints.
	
	Note that some decision variables in model \eqref{prob:ILP} can be eliminated. E.g., the variable $\alpha$ can be totally removed by formulation \eqref{cons:2} as $$\alpha_k = \delta_k-\sum_{i=0}^{k-2}\sum_{j=1}^{k-1}\gamma_{ijk}, \forall k \in\IntEnt{1,n}.$$
	Anyway, despite the above simplification, the use of trigram model requires amounts of $O(n^3)$ binary variables in $\gamma$ whose order is not reducible. 
	
	We observe that there are very few information (only in \ref{cons:7} for PP or SBAR) about syntactic structures of sentences, thus it is highly possible to generate sentences ungrammatically correct. Clarke introduced some compression-specific constraints such as the modifier constraints, argument structure constraints and discourse constraints for the propose of some basic linguistical and semantical checkings. However, deletions (or non-deletions) are totally built on the probability of words which yields an amplification of the defect of probability based models. Furthermore, some constraints for grammar checking are limited by different sentence types. Although the sentence is grammatically correct, it is possible that the compressed sentence isn't syntactically acceptable. Also, introducing too many constraints to check the grammar yields ILP problem more difficult to be solved and even intractable for complex sentences.
	
	Different to Clarke's work, we propose to combine the parse tree model for linguistical and semantical checkings.

\subsection{Parse Tree Model}\label{subsec:parsetreemodel}

A parse tree is an ordered, rooted tree which reflects the syntax of the input language according to some grammar rules (e.g. syntax-free grammar (CFG)). The interior nodes are labeled by nonterminal categories of the grammar, while the leaf nodes are labeled by terminal categories. E.g., a statement ``The man saw the dog with the telescope'' is parsed as follows:
\begin{figure}[htbp!]
	\centering 
	\includegraphics[width=10cm]{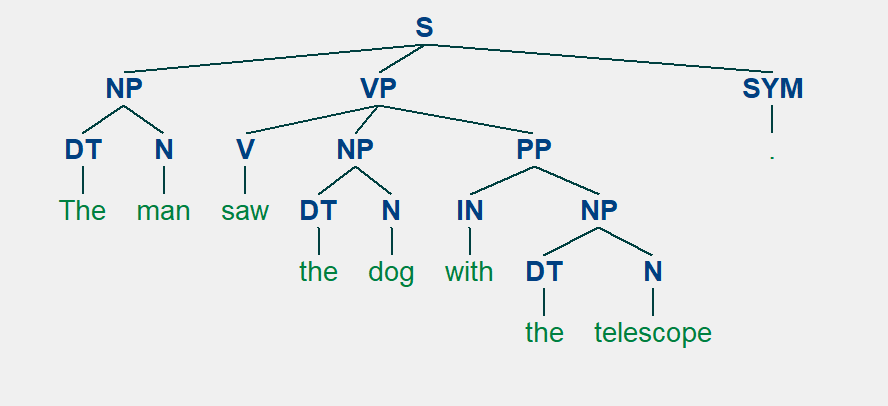}
	\caption{Parse tree example}
	\label{fig:1}
\end{figure}
\begin{itemize}
	\item
	
	S for sentence, the top-level structure.
	\item
	
	NP for noun phrase.
	\item
	
	VP for verb phrase.
	\item
	
	SYM for punctuation.
	
	\item
	
	N for noun.
	
	\item
	
	V for verb. In this case, it is the transitive verb `saw'.
	\item
	
	DT for determiner, in this instance, it is the definite article `the'.
	
	\item PP for preposition phrase.
	\item
	IN for preposition, which often starts a complement or an adverbial.
\end{itemize}
The reader can refer to Appendix \ref{appendix:A} for full list of supported in our model.

The parse tree is constructed in three steps: Tokenization, POS Tagging and Parsing. \textit{Tokenization} will separate the sentence into words and punctuations; \textit{POS Tagging} will tag the part-of-speech of each word; \textit{Parsing} will build a parse tree based on some grammar rules. For constructing a parse tree in practice, we can use NLTK in Python or Stanford CoreNLP package \cite{Software_StanfordCoreNLP,paper_manning_2014} in Java. Our example in Figure \ref{fig:1} is built with NLTK by defining syntax-free grammar (CFG) and using a Recursive Descent Parser. For more information about CFG and Recursive Descent Parser, the reader can refer to \cite{book_steven_2009}.

As observed in Figure \ref{fig:1}, a higher level always indicates a more important component, whereas a lower level tends to carry more semantic content. To analyze the sentence structure with a parse tree, the top-level (or zero level) is the node S for the whole sentence. At the first level, the node NP contains the subject of the sentence, the node VP contains the predicate and the node SYM stands for punctuation. Further analyzing, the VP part contains the main verb part V and the object of the sentence NP followed by a preposition phrase PP which infers to the adverbial phrase. Therefore, a parse tree presents the clear syntactic structure of a sentence in a logical way. Taking this advantage, we can use it to handle the grammar checking. 

With the parse tree, the sentence compression task can be considered as cropping the parse tree to find a subtree remaining grammatically correctness and containing main meaning of the source sentence. E.g., the sentence in Figure \ref{fig:1} can be compressed by deleting the adverbial ``with the telescope" (i.e., deleting the node PP).

\begin{figure}[htbp!]
	\centering
	\begin{minipage}[c]{0.5\textwidth}
		\centering
		\includegraphics[height=3.5cm]{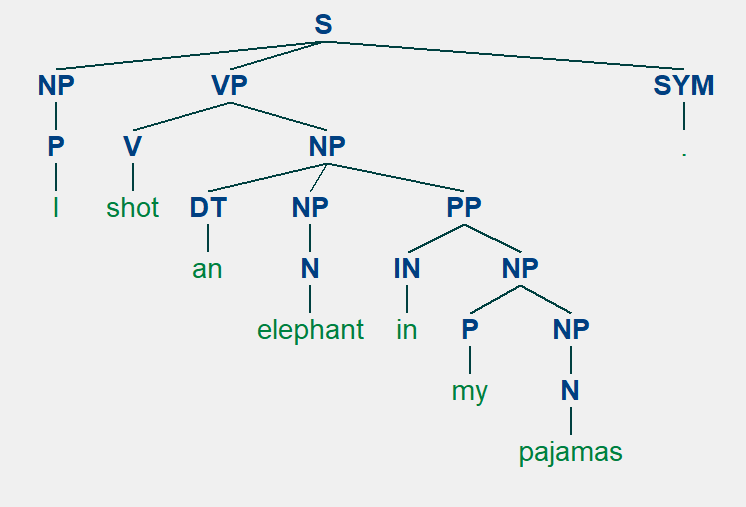}
	\end{minipage}%
	\begin{minipage}[c]{0.5\textwidth}
		\centering
		\includegraphics[height=3.5cm]{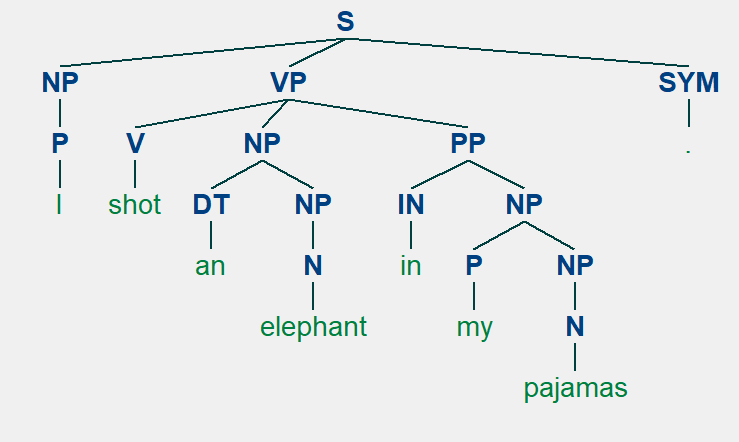}
	\end{minipage}%
	\caption{Ambiguity in sentence structure}
	\label{fig:2}
\end{figure}

However, due to the existence of ambiguity, one sentence can be parsed in many ways once the coverage of the grammar increases and the length of the input sentence grows. A well-known example of ambiguity is given in the Groucho Marx movie, Animal Crackers (1930) as: ``I shot an elephant in my pajamas.'' Using the following CFG grammar:
\begin{itemize}
	\item[] S $\to$ NP VP SYM
	\item[] PP $\to$ IN NP
	\item[] NP $\to$ DT NP $|$ DT NP PP $|$ P $|$ N $|$ P NP
	\item[] VP $\to$ V NP $|$ V NP PP
	\item[] DT $\to$ ``an"
	\item[] N $\to$ ``elephant" $|$ ``pajamas"
	\item[] V $\to$ ``shot"
	\item[] P $\to$ ``in"
	\item[] P $\to$ ``my"
\end{itemize}
this sentence can be analyzed in two ways, depending on whether the prepositional phrase \emph{in my pajamas} describes the elephant or the shooting event. Two different parse trees are illustrated in Figure \ref{fig:2}.

Therefore, the development of a fiable CFG grammar is very important to the parse tree model. To this end, we develop a CFG grammar generator which helps to generate automatically a CFG grammar based on the target sentence and fundamental structures of different phrase types. Then, a recursive descent parser can help to build parse trees. 

\subsection{Hybrid Model (ILP-Parse Tree Model)}

Our proposed model for sentence compression, namely ILP-Parse Tree Model (ILP-PT), is based on the combination of the two models described above. The ILP model will provide some candidates for compression with maximal probability, while the parse tree model helps to guarantee the grammar rules and keep the main meaning of the sentence. Our model reads in four steps:

\noindent\textbf{Step 1 (Build ILP model):} as in formulation (\ref{prob:ILP}).
	
\noindent\textbf{Step 2 (Parse sentence):} as described in subsection \ref{subsec:parsetreemodel}.

\noindent\textbf{Step 3 (Fix variables for sentence trunk):} Identifying the sentence trunk in the parse tree and fixing the corresponding integer variables to be $1$ in ILP model. This step helps to extract the sentence trunk by keeping the main meaning of the original sentence while reducing the number of binary decision variables. 

More precisely, we will introduce for each node $N_i$ of the parse tree a label $s_{N_i}$ taking the values in $\left\{0, 1, 2\right\}$. A value $0$ represents the deletion of the node; $1$ represents the reservation of the node; whereas $2$ indicates that the node can either be deleted or be reserved. We set these labels as \emph{compression rules} for each CFG grammar to support any sentence type of any language. 

\begin{figure}[htbp!]
	\centering
	\includegraphics[width=10cm]{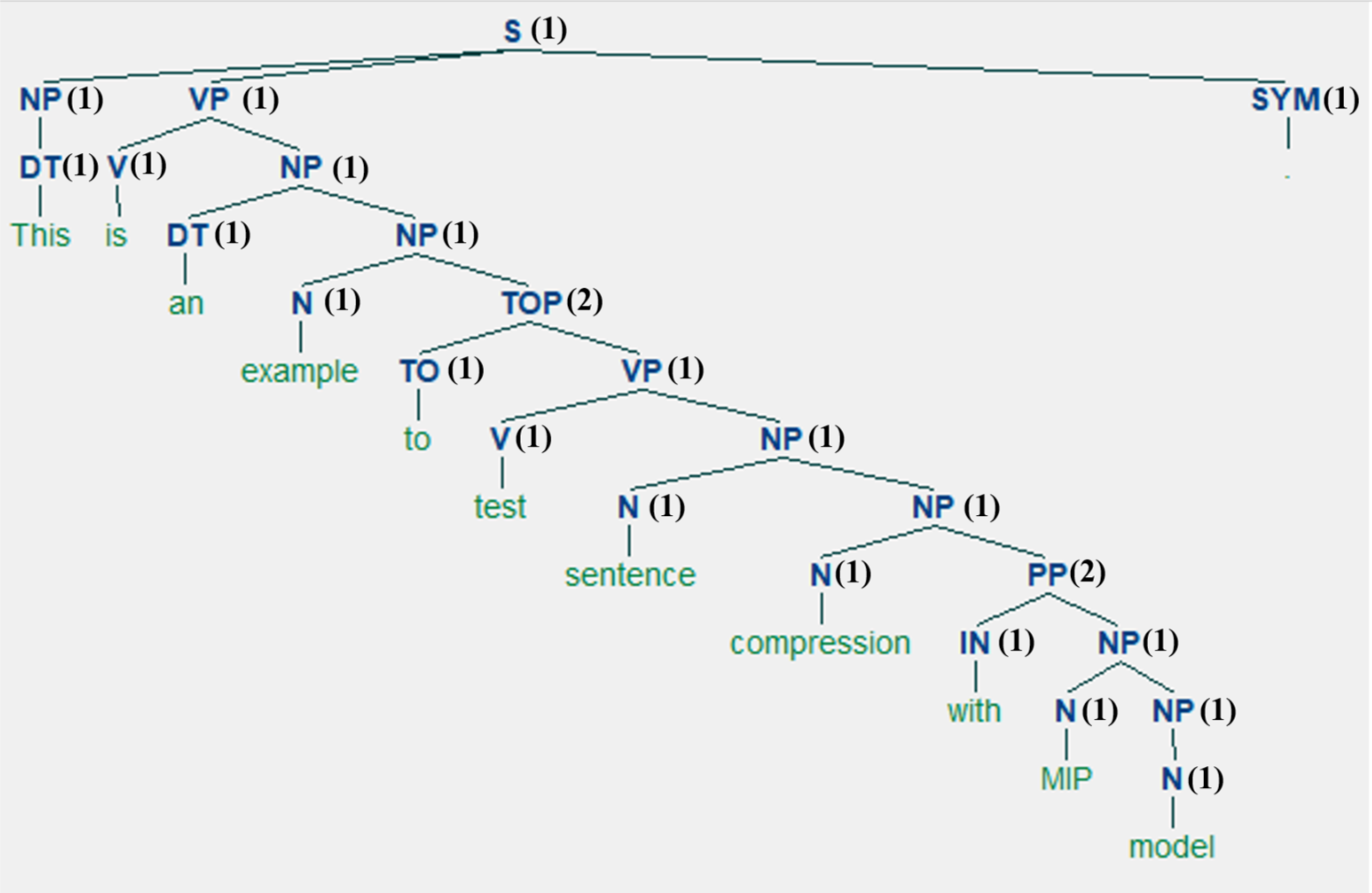}
	\caption{Parse tree compression}\label{fig:3}
\end{figure}

As an example, the statement ``This is an example to test sentence compression with MIP model." is parsed in Figure \ref{fig:3}.
For each word $x_i$, we go through all its parent nodes till the root S. If the traversal path contains $0$, then $\delta_i=0$; else if the traversal path contains only $1$, then $\delta_i=1$; otherwise $\delta_i$ will be further determined by solving the ILP model. The sentence truck is composed by the words $x_i$ whose $\delta_i$ are fixed to $1$. Using this method, we can extract the sentence trunk and reduce the number of binary variables in ILP model. As a result, we get in Figure \ref{fig:3} the sentence truck as : ``This is an example to test sentence compression."

Appendix \ref{appendix:B} lists some default compression rules that we can use to decide the elimination of each node. Of course, the supported sentence structures and the corresponding compression rules can be enriched by users to support any other sentence type of any language.

\noindent\textbf{Step 4 (Solve ILP):} Applying an ILP solution algorithm to solve the simplified ILP model derived in Step 3 and generate a compression. In the next section, we will focus on DC programming approach for solving ILP.

\section{PDCABB Algorithm for ILP}\label{sec:solvethemodel}
ILP model can be stated in a standard matrix form as: 
\begin{equation*}\label{P}
\min \{f(x):=c^{\top}x ~|~ x\in \mathcal{S}\}
\tag{$P$},
\end{equation*}
where $c\in \R^n$, $b \in \R^m$, $A \in \R^{m\times n}$, and the set $\mathcal{S} := \{x\in \{0,1\}^n ~|~ Ax = b\}$ is assumed to be nonempty. The linear relaxation of the set $\mathcal{S}$ is denoted by $\mathcal{K}$ defined as $\{x\in [0,1]^n ~|~ Ax = b\}$. Clearly, $\mathcal{S} = \mathcal{K} \cap \{0,1\}^n.$
Solving an ILP is in general NP-hard. A most frequently used method for ILP is the branch-and-bound algorithm. Gurobi \cite{software_gurobi} is one of the best ILP solvers which is an efficient implementation of the branch-and-bound combining various techniques such as presolving, cutting planes, heuristics and parallelism etc.

In this section, we will present a hybrid approach, namely Parallel-DCA-Branch-and-Bound (PDCABB) algorithm, for solving problem \eqref{P}. The non-parallel DCABB algorithm for binary linear program is introduced in \cite{LeThi2001} by Le Thi and Pham, then the general cases with mixed-integer linear/nonlinear optimization are extensively developed by Niu and Pham (see e.g., \cite{paper_niu_2008,thesis_niu_2010,paper_niu_2011}) where the integer variables are not supposed to be binary. There are various applications of this kind of approaches including scheduling \cite{paper_lethi_2009a}, network optimization \cite{paper_schleich_2012}, cryptography \cite{paper_lethi_2009c} and finance \cite{paper_lethi_2009b,paper_pham_2016} etc. This kind of algorithm is based on continuous representation techniques for integer set, exact penalty theorem, DC algorithm (DCA), and Branch-and-Bound scheme. Recently, we have developed a parallel version of DCABB (called PDCABB) \cite{PDCABB,paper_niu_2018} which uses the power of multiple CPUs and GPUs for accelerating the convergence of DCABB. Next, we will present respectively DCA, DCABB and PDCABB.

\subsection{DC Program and DCA}
Let $\R^n$ denote the $n$-dimensional Euclidean space equipped with the standard inner product $\langle \cdot, \cdot \rangle$ and the induced norm $\Vert \cdot \Vert$. The set of all proper, closed and convex functions of form $\R^n\to (-\infty,+\infty]$ is denoted by $\Gamma_0(\R^n)$, which is a convex cone; and $\text{DC}(\R^n):=\Gamma_0(\R^n) - \Gamma_0(\R^n)$ is the set of DC (Difference-of-Convex) functions (under the convention that $(+\infty) - (+\infty) = +\infty$), which is in fact a vector space spanned by $\Gamma_0(\R^n)$. If $f=g -h$ with $(g, h) \in \Gamma_0(\R^n)^2$, then $g$ and $h$ are called DC components of $f$.

The standard DC program is given by  
\begin{equation}
\label{primal1}
\alpha = \text{min}\; \{\phi(x):= \phi_1(x) - \phi_2(x) ~|~ x\in\R^n\}, \tag{$P_{dc}$}
\end{equation}
where $\phi_1, \phi_2 \in \Gamma_0(\R^n)$ and $\alpha$ is assumed to be finite, this implies that $\emptyset \neq \dom \phi_1 \subset \dom \phi_2$. A point $x^*$  is called a \textit{critical point} of problem (\ref{primal1}) if $\partial \phi_1(x^*)  \cap \partial \phi_2(x^*) \neq \emptyset$.
A well-known DC Algorithm, namely DCA, for \eqref{primal1} was firstly introduced by D.T. Pham in 1985 as an extension of the subgradient method, and extensively developed by H.A. Le Thi and D.T. Pham since $1994$. DCA consists of solving the standard DC program by a sequence of convex ones as
$$x^{k+1}\in \argmin \{ \phi_1(x) - \langle x, y^k \rangle ~|~ x\in \R^n \}$$
with $y^k\in \partial \phi_2(x^k)$. This convex program is derived by convex-overestimating the DC function $\phi$ at iterate $x^k$, denoted by $\phi^k$, which can be constructed by linearizing the component $\phi_2$ at $x^k$ and taking $y^k\in \partial \phi_2(x^k)$ verifying:
$$
\phi(x) = \phi_1(x)-\phi_2(x)\leq \phi_1(x) - (\phi_2(x^k)+\langle x-x^k, y^k \rangle)= \phi^k(x), \forall x\in \R^n.
$$ 
The detailed DCA for standard DC program \eqref{primal1} reads:
\begin{algorithm}[h!]
	\caption{DCA}
	\label{algo:DCA0}
	
	\textbf{Initialization:} Choose $x^0 \in \dom(\partial \phi_2)$.
	
	\textbf{Iterations:} For $k = 0,1,2,\dots$
	
	Compute $y^k \in \partial \phi_2(x^k)$.
	
	Find $
	x^{k+1} \in \argmin\{ \phi_1(x)-\langle x,y^k \rangle ~|~ x\in\R^n\}.$	
\end{algorithm}
\begin{theorem}[Convergence of DCA \cite{Pham1997,LeThi2005}]
	\label{thm:1}
	Let $\{x^k\}$ and $\{y^k\}$ be bounded sequences generated by DCA (Algorithm \ref{algo:DCA0}) for problem \eqref{primal1}, then
	\begin{enumerate}
		\item The sequence $\{f(x^k)\}$ is decreasing and bounded below.
		\item Any limit point of $\{x^k\}$ is a critical point of problem \eqref{primal1}.
		\item If either $\phi_1$ or $\phi_2$ is polyhedral, then problem $\eqref{primal1}$ is called polyhedral DC program, and DCA is terminated in finitely many iterations.
	\end{enumerate}
\end{theorem} 
For more results on DC program and DCA, the readers can refer to  \cite{Pham1997,Tao1998,LeThi2005,LeThi2018,paper_niu_2018a} and the references therein.

\subsection{DC Formulation and DCA for Problem \eqref{P}}
We will show that problem \eqref{P} can be equivalently represented as a standard DC program in form of \eqref{primal1}. Using the \emph{continuous representation technique}, we can reformulate the binary set $\{0,1\}^n$ as a set involving continuous variables and continuous functions only. A classical way is using a nonnegative finite function $p:[0,1]^n\mapsto \R_+$ to rewrite the binary set as:
$$\{0,1\}^n = \{x\in [0,1]^n~|~p(x)\leq 0\}.$$
Then $$\mathcal{S} = \{(x,y)\in \mathcal{K}~|~ p(x)\leq 0\}.$$
There are many alternative functions for $p$, some frequently used functions are listed in Table \ref{table:1}, where $p_1$ is a concave piecewise linear function, $p_2$ is a concave quadratic function, and $p_3$ is a trigonometric function. 
\begin{table}[h!]
	\centering
	\caption{Alternative functions for $p$}
	\label{table:1}
	\begin{tabular}[h]{c|l|c}
		\hline
		denotation of $p$ & expression of $p$ & DC components of $p$\\
		\hline
		$p_1$ & $\sum_{i=1}^{n} \min\{x_i,1-x_i\}$ & $g_1(x) = 0$, $h_1(x)=-p_2(x)$ \\
		\hline
		$p_2$ & $\sum_{i=1}^{n}x_i(1-x_i)$ & $g_2(x) = 0$, $h_2(x)=-p_2(x)$\\
		\hline
		$p_3$ & $\sum_{i=1}^{n}\sin^2(\pi x_i)$ & $g_3(x) = \pi^2 \|x\|^2$, $h_3(x) = g_3(x) - p_3(x)$\\
		\hline
	\end{tabular}
\end{table}

Let us define the penalized problem as follows: 
\begin{equation}
\label{Pt}
\alpha_t =  \min \{c^{\top}x + t p(x) ~|~ x \in \mathcal{K}\}.
\tag{$P_t$}
\end{equation}
Thanks to the concavity and nonnegativity of $p_1$ and $p_2$, we have the exact penalty theorem \cite{LeThi_1999,LeThi_2012} as follows:
\begin{theorem}[Exact Penalty Theorem \cite{LeThi_1999,LeThi_2012}]
	\label{thm:exactpenalty}
	For $p=p_1$ or $p_2$, there exists a finite number $t_0 \geq 0$ such that for all $t > t_0$, problem \eqref{P} is equivalent to problem \eqref{Pt}.
\end{theorem}
The equivalence means that problems \eqref{P}
and \eqref{Pt} have the same set of global optimal solutions. The penalty parameter $t_0$ can be computed by: 
$$t_0 = \frac{\min\{c^{\top}x ~|~ x\in \mathcal{S}\} -\alpha_0}{m},$$ 
where $m = \min \{p(x)~|~ x\in V(\mathcal{K}), p(x) > 0\}$ under the convention that $m=+\infty$ if $\{x\in V(\mathcal{K})~|~p(x) > 0\} = \emptyset$ and $\frac{1}{+\infty} = 0$. In practice, computing $t_0$ is difficult since both $\min\{c^{\top}x ~|~x\in \mathcal{S}\}$ and $\min \{p(x)~|~ x\in V(\mathcal{K}), p(x) > 0\}$ are nonconvex optimization problems which are difficult to be solved, while the computation of $\alpha_0$ is easy which requires solving a linear program.

However, for $p=p_3$, the penalization in \eqref{Pt} is inexact. In this case, it is well-known that for $\{x_t^*\}_{t\geq 0}$ being a sequence of optimal solution of \eqref{Pt}, all the converging sub-sequences extracted from $\{x_t^*\}_{t\geq 0}$ converge to optimal solutions of problem \eqref{P}, see e.g., \cite{DP_book}.

Note that for exact penalty, the computation of $t_0$ is difficult especially due to the term $m$. To estimate a suitable $t_0$, one may find an upper bound of $t_0$ by estimating an upper bound of $\min\{c^{\top}x ~|~ x\in \mathcal{S}\}$ and a nonzero lower bound of $m$. An upper bound of $\min\{c^{\top}x ~|~ x\in \mathcal{S}\}$ is not too difficult to be obtained since $\min\{c^{\top}x ~|~ x\in \mathcal{S}\}\leq c^{\top} \tilde{x}, \forall \tilde{x}\in \mathcal{S}$. However, the term $m=\min \{p(x)~|~ x\in V(\mathcal{K}), p(x) > 0\}$ (and its lower bound) is hard to be computed if $V(\mathcal{K})$ is unknown. Note that $m$ depends only on $V(\mathcal{K})$, and it is possible to have a vertex $y\in V(\mathcal{K})$ such that $p(y)>0$ but very close to $0$, thus $m$ could be very small. It is an interesting question to investigate how to estimate a nonzero lower bound of $m$ over a particular structured polyhedral set $\mathcal{K}$ for sentence compression in our future work.

In practice, instead of computing $t_0$, we can increase the parameter $t$ and check whether the computed solution of problem \eqref{Pt} is changed or not. Based on Theorem \ref{thm:exactpenalty}, the computed solution will not be changed for large enough $t$. In our numerical test, to simplify the computation of $t$, we will fix $t$ as a large number, e.g., $t=10^5$ is suitable for all tests of sentence compression with $n\leq 100$ words.

By introducing the indicator function $\chi_{\mathcal{K}}$ with $\chi_{\mathcal{K}}(x)$ equals to $0$ if $x\in \mathcal{K}$ and $+\infty$ otherwise, and using the DC decompositions of $p$ given in Table \ref{table:1}, we can rewrite problem (\ref{Pt}) as the following standard DC program:
\begin{equation}\label{$DCP$}
\min \{F^t(x):= \underbrace{(\chi_{\mathcal{K}}(x) + tg(x))}_{g^t(x)}- \underbrace{(-c^{\top}x + th(x))}_{h^t(x)} ~|~ x\in \R^n \}, \tag{$P_{dc}^t$}
\end{equation}
Clearly, the objective function $F^t$ is a DC function, and the above formulation is a standard DC program. 

To apply DCA for problem (\ref{$DCP$}), we need :
\begin{enumerate}
	\item[(i)] Compute a sub-gradient of $h^t$ at iterate $x^k$ by
	\begin{equation}
	\label{eq:dh}
	y^k \in \partial h^t(x^k) = -c + t u, \text{ with } u \in \partial h(x^k),
	\end{equation}
	where the expressions of $\partial h(x^k)$ is given in Table \ref{table:2} whose calculations are easy and fundamental in convex analysis \cite{book_rockafellar_1970}.
	\begin{table}[h!]
		\centering
		\caption{ Expressions of $\partial h (x)$}
		\label{table:2}
		\begin{tabular}[h]{c|c}
			\hline
			denotation of $\partial h(x)$ & expression of $\partial h(x)$ \\
			\hline
			$\partial h_1 (x)$ & $\left\{ u\in \R^n ~\big|~ u_i \in \begin{cases}
			\{1\} & \text{if } x_i >\frac{1}{2}\\
			[-1,1] & \text{if } x_i = \frac{1}{2}\\
			\{-1\} & \text{if } x_i < \frac{1}{2}
			\end{cases}, i\in \IntEnt{1,n} \right\}$ \\
			\hline
			$\partial h_2 (x)$ & $\{2x-1\}$ \\
			\hline
			$\partial h_3 (x)$ & $\{2\pi^2 x - \pi \sin(2\pi x)\}$\\
			\hline
		\end{tabular}
	\end{table}
	\item[(ii)] The next iterate $x^{k+1}$ is computed by solving the optimization problem
	$$x^{k+1} \in \argmin \{g^t(x) -\langle y^k, x \rangle ~|~ x\in \mathcal{K} \}$$
	which is a linear program for $g=g_i$ or $g_2$, and convex quadratic program for $g=g_3$.
	\item[(iii)] DCA could be terminated if $\|x^{k+1} - x^k\|$  or $\|F^t(x^{k+1}) - F^t(x^k)\|$ is smaller than a given tolerance. 
\end{enumerate}
The detailed DCA for problem (\ref{$DCP$}) is summarized in Algorithm \ref{algo:DCA} with the same convergence theorem stated in Theorem \ref{thm:1}.
\begin{algorithm}[h!]
	\caption{DCA for (\ref{$DCP$})}
	\label{algo:DCA}
	\KwIn{Initial point $x^0\in \R^n$; large penalty parameter $t>0$; small tolerances $\varepsilon_1>0$ and $\varepsilon_2>0$.}
	\KwOut{Computed solution $x^*$ and objective value $F^*$;} 
	
	\textbf{Initialization:} Set $k \leftarrow 0$. 
	
	\textbf{Step 1:} Compute $y^k \in \partial h^t(x^k)$ via formulations \eqref{eq:dh} and expressions in Table \ref{table:2};
	
	\textbf{Step 2:} Solve problem $x^{k+1} \in \argmin \{ g^t(x) - \langle x, y^k \rangle ~|~ x \in K \}$ via expressions in Table \ref{table:1} ;
	
	\textbf{Step 3:} (Stopping criteria) 	
	
	\SetKwIF{If}{ElseIf}{Else}{if}{then}{else if}{else}{end}
	\eIf{$\|x^{k+1}-x^{k}\| \leq \varepsilon_1$ or $|F^t(x^{k+1})-F^t(x^{k})| \leq \varepsilon_2$}
	{
		$x^* \leftarrow x^{k+1}$; $F^* \leftarrow F_i^t(x^{k+1})$; \Return;
	}
	{
		$k\leftarrow k+1$; \textbf{Goto} Step 1.
	}
\end{algorithm}

Concerning on the choice of the penalty parameter $t$, as far as we know, there is still no explicit formulation to compute a suitable parameter. The exact penalty theorem  only proved the existence of such a penalty parameter, but it is hard to obtain the required $t_0$. In practice, we suggest using the following two methods to handle this parameter: the first method is to take arbitrarily a large positive value for parameter $t$; the second one is to increase $t$ by some ways in iterations of DCA (see e.g., \cite{thesis_niu_2010,paper_pham_2016}). Note that a small but large enough parameter $t$ is preferable in computation, since the problem will become ill-conditioned and slow down (maybe destroy) the convergence of DCA in practice if $t$ is set too large.

\subsection{Parallel-DCA-Branch-and-Bound Algorithm}
DCA can often terminate very quickly to provide feasible solutions to problem (\ref{P}). Therefore, it is proposed as a good upper bound solver for problem \eqref{P} by introducing DCA into a parallel-branch-and-bound framework.

In this subsection, we briefly introduce the Parallel-DCA-Branch-and-Bound (PDCABB) algorithm proposed in \cite{paper_niu_2018} for solving problem \eqref{P}. Let us denote the linear relaxation problem of (\ref{P}) as \ref{RP} defined by:  
\begin{equation*}\label{RP}
\min \{f(x) ~|~ x\in \mathcal{K}\}, \tag*{$R(P)$}
\end{equation*}
whose optimal value, denoted by $l(P)$, is a lower bound of (\ref{P}). The Branch-and-Bound procedure is similar to the classical one which consists of two blocks: \textit{Root Block} and \textit{Node Block}. 

\subsubsection{Root Block}
The root block consists of solving the linear relaxation \eqref{RP} and check the feasibility of the lower bound solution, if the solution is feasible, then we terminate the algorithm; otherwise, we will run DCA in parallel with randomly generated initial points in $[0,1]^n$. Suppose that we have $s$ ($s\geq 1$) available CPU cores, then we can run $s$ DCA simultaneously. We can update the upper bound solution as the best feasible solution obtained by DCA. The Root operations are terminated by creating a list of nodes $L$ with problem \eqref{P} as the initial node in $L$.

\subsubsection{Node Block}
The node block consists of a loop by processing each nodes in the list $L$. We can select (the node selections will be presented later) and remove a sublist of nodes (between $1$ to $s$) from $L$, denoted by $L_s$, then we can process these nodes in parallel as follows: for each node problem $(P_i)\in L_s$, we start by solving $R(P_i)$ to get a lower bound solution $x^*$ on the node problem $(P_i)$. If $x^*$ is better than the upper bound solution, then we update the upper bound; Else if the gap between the upper bound and $l(P_i)$ is larger than a given tolerance $\varepsilon_3$, then there is likely a better feasible solution for $(P_i)$, thus we suggest restart DCA from $x^k$ for DC formulation of $(P_i)$, and check whether we indeed find a better feasible solution to update upper bound. Otherwise, we will make branches from the node $(P_i)$ based on some \textit{branching rules} if and only if the gap between upper bound and $l(P_i)$ is not smaller than a tolerance $\varepsilon_4$. Some available branching rules will be given later. Once all nodes in $L_s$ have been explored, we will add all new created nodes in the list $L$ and repeat the loop until the list $L$ is empty.

The detailed PDCABB is summarized in Algorithm \ref{algo:PDCABB}. Note that there are two tolerances used in PDCABB. The tolerance $\varepsilon_3$ is  for restarting DCA, and the tolerance $\varepsilon_4$ is for checking the gap between upper and lower bound (i.e., $\varepsilon_4$-optimality of the computed solution). 

\subsubsection{Node Selections}
We use the following optional strategies to select parallel nodes:
\begin{itemize}
	\item[$\bullet$] \textbf{Best-bound search}: choose active nodes with the best objective values in their associated linear relaxations;
	\item[$\bullet$] \textbf{Depth-first search}: select the most recently created nodes;
\end{itemize} 

\subsubsection{Branching Strategies}
Once a lower bound solution $x^*$ is obtained by solving a linear relaxation $R(P_i)$, we propose to use the following optional methods to find an index $j$ with $x^*_{j}\notin\{0,1\}$, and two new branches are established from the node $(P_i)$ by adding the constraint $x_j=0$ and $x_{j}=1$ respectively into $(P_i)$. Let $J = \{k\in \IntEnt{1,n} ~|~ x_k^* \notin \{0,1\}\}$, one can 
\begin{enumerate}
	\item select $j \in \argmin_{k\in J} \vert x_k^* - \frac{1}{2} \vert$;
	\item select $j \in \argmax_{k\in J} \min\{x_k^*,1-x_k^*\}$;
	\item select $j \in \argmax_{k\in J}\vert c_k \vert$.
\end{enumerate}
The reader can refer to \cite{paper_niu_2018,paper_niu_2008,thesis_niu_2010} for more discussions about this algorithm.

\begin{algorithm}[h!]
	\caption{PDCABB}
	\label{algo:PDCABB}
	\KwIn{Problem (\ref{P}); number of parallel workers $s$; tolerances $\varepsilon_3>0$ and $\varepsilon_4>0$;}
	\KwOut{Optimal solution $x_{opt}$ and optimal value $f_{opt}$;} 
	
	\textbf{Initialization:} $x_{opt}=null$; $f_{opt}=+\infty$.
	
	\textbf{Step 1: Root Block}
	
	Solve $R(P)$ to obtain its optimal solution $x^*$ and optimal value $l(P)$;
	
	\SetKwIF{If}{ElseIf}{Else}{if}{then}{elseif}{else}{end}
	\uIf{$R(P)$ is infeasible}
	{
		
		\Return the problem is infeasible;
	}
	\ElseIf{$x^*\in S$}
	{
		$x_{opt} \leftarrow x^*$; $f_{opt} \leftarrow l(P)$; \Return;
	}
	
	\SetKwFor{For}{parallelfor}{do}{end}   
	\For{$i=1,\ldots,s$}
	{
	Run DCA for (\ref{Pt}) from random initial point to get $\bar{x}^*$;
	
	\If{$\bar{x}^*\in S$ and $f(\bar{x}^*)<f_{opt}$}
	{Update best upper bound solution $x_{opt} \leftarrow \bar{x}^*$ and $f_{opt} \leftarrow f(\bar{x}^*)$;
	}
	}

	$L \leftarrow \{P\}$;
	
	\textbf{Step 2: Node Block}
	
	\While{$L\neq \emptyset$}
	{
		Select a sublist $L_s$ of $L$ with at most $s$ nodes in $L_s$;
		
		Update $L\leftarrow L \setminus L_s$;
		
		\SetKwFor{For}{parallelfor}{do}{end}   
		\For{$P_i \in L_s$}
		{
			Solve $R(P_i)$ and get its solution $x^*$ and lower bound $l(P_i)$;		
			
			\SetKwIF{If}{ElseIf}{Else}{if}{then}{elseif}{else}{end}
			\If{$R(P_i)$ is feasible and $l(P_i) < f_{opt}$}
			{
				\uIf {$x^* \in S$}
				{
					$x_{opt}\leftarrow x^*$; $f_{opt} \leftarrow l(P_i)$; 
				}
				\Else
				{
					\uIf {$f_{opt}-l(P_i)> \varepsilon_3$}
					{
						Restart DCA for $(P_i^t)$ from $x^*$ to get its solution $\hat{x}^*$;
						
						\If {$\hat{x}^* \in S$ and $f_{opt} > f(\hat{x}^*)$}
						{
							$x_{opt}\leftarrow \hat{x}^*$; $f_{opt} \leftarrow f(\hat{x}^*)$; 
							
						}
					}
					\ElseIf{$f_{opt}-l(P_i) > \varepsilon_4$}
					{
						Branch $P_i$ into two new problems $P_i^u$ and $P_i^d$;
						
						Update $L\leftarrow \{P_i^u , P_i^d\}$;
					}
				}
			}
		}
	}
\end{algorithm}

\section{Experimental Results}\label{sec:experimental_result}
In this section, we present some experimental results for assessing the quality of the hybrid ILP-PT model and the performance of PDCABB algorithm. 

Our sentence compression model is implemented in Python as a component of the Natural Language Processing package, namely \verb|NLPTOOL| (actually supporting multi-language tokenization, tagging, parsing, automatic CFG grammar generation, and sentence compression). We use NLTK 3.2.5 \cite{NLTK} for creating parsing trees and Gurobi 7.5.2 \cite{software_gurobi} for solving the linear programs $R(P_i)$ required in DCA. We use an implementation of PDCABB algorithm \cite{PDCABB} coded in C++ which provides a Python interface.

\subsection{Corpora}

We use three public available corpora in our experiments. The Brown Corpus is used for training POS tagger. This corpus is the first computer-readable general corpus for linguistic research on modern English and well supported by NLTK.

Two corpora are used for computing probabilities based on $N$-gram model: the Penn Treebank corpus for general propose (namely \textit{Treebank}, provided in NLTK), and a mix corpus designed for sentence compression only, namely \textit{Clarke+Google}, which are collected from the Clarke's written corpus (used in \cite{paper_clarke_2008}) and Google's corpus (used in \cite{paper_filippova_2013}). The Kneser-Ney smoothing is used for the probabilities of missing words.

\subsection{F-score Evaluation}

We use a statistical approach called \emph{F-score} to evaluate the similarity between the compression computed by our algorithm and a standard compression provided by human. F-score is defined by : $$F_{\mu} = (\mu^2+1)\times \frac{P\times R}{\mu^2 \times P + R}$$
	where $P$ and $R$ represent for precision rate and recall rate as:
	$$P = \frac{A}{A+C}, R = \frac{A}{A+B}$$
	in which $A$ denotes for the number of words both in the compressed result and the standard result; $B$ is the number of words in the standard result but not in the compressed result; and $C$ counts the number of words in the compressed result but not in the standard result. The parameter $\mu\geq 0$, called preference parameter, stands for the preference between precision rate and recall rate for evaluating the quality of the results. $F_{\mu}$ is a strictly monotonic function defined on $[0,+\infty[$ with $\displaystyle\lim_{\mu\to 0} F_{\mu}=P$ and $\displaystyle\lim_{\mu\to +\infty} F_{\mu}=R$. In our tests, we will use $F_1$ as F-score. Clearly, a larger F-score indicates a better compression.
	
	E.g., given an original sentence ``The aim is to give councils some control over the future growth of second homes." and a standard compression (human compression) ``The aim is to give councils control over the growth of homes.". We compute F-score for a compression ``aim is to give councils some control." as $A=7,B=6,C = 1$, then $P=87.5 \%, R=53.8\%$, and F-score is $66.7 \%$.

\subsection{Numerical Results}

Table \ref{tab:results1} illustrates the compression results obtained by Clarke-Lapata ILP probabilistic model (P) and our hybrid ILP-PT model (H) for 285 sentences randomly chosen in a compression corpus \cite{paper_filippova_2013}. The compression rates are given by $50 \%$, $70 \%$ and $ 90 \%$ respectively. We compare the average solution time and the average F-score for these models solved by Gurobi and PDCABB for different models and corpora. The experiments are performed on a laptop equipped with Intel i5-4200H 2.80GHz CPU (2 cores) and 8GB RAM. It can be observed that our hybrid model often provides better F-scores in average for all compression rates, while the computing time for both Gurobi and PDCABB are all very short within less than 0.15 seconds. We can also see that Gurobi and PDCABB provided slightly different solutions in F-score since the global optimal solutions may not be unique in general and the branch-and-bound based algorithm often provides $\varepsilon$-optimal solutions with gap between the upper and lower bounds being smaller than the tolerance $\varepsilon$ (here we use $\varepsilon = 10^{-5}$ in our experiments). The reliability of our judgment based on F-scores is trustworthy since two different algorithms provide very similar F-scores on same models and same sentences.
\begin{table}[h!]
	\caption{Compression results}\label{tab:results1}
	\resizebox{330pt}{48pt}{
		\begin{tabular}[h]{|c|c|c|c|c|c|c|c|} 
			\hline
			\multirow{2}*{Corpus+Model} & \multirow{2}*{Solver} & \multicolumn{2}{c|}{$50 \%$ CR} & \multicolumn{2}{c|}{$70 \%$ CR} & \multicolumn{2}{c|}{$90 \%$ CR} \\
			\cline{3-8}
			& & F-score (\%) & Time(s) & F-score(\%) &  Time(s) &  F-score(\%) &  Time(s)\\
			\hline
			\multirow{2}*{Treebank+P} & Gurobi & 61.14  & 0.051 & 74.04 & 0.057 & \bfseries 80.83 & 0.034 \\
			\cline{2-8}
			& PDCABB &  \bfseries 62.24 & 0.104 &{\bfseries 74.42} & 0.092 & 80.25 & 0.055\\
			\hline
			\multirow{2}*{Treebank+H} & Gurobi & \bfseries 82.11 & 0.025 & \bfseries 82.37 & 0.030 &  82.48 & 0.022\\
			\cline{2-8}
			& PDCABB &82.10 & 0.030 & 82.34 & 0.048 & \bfseries 82.49 &0.037\\
			\hline
			\multirow{2}*{Clarke+Google+P} & Gurobi & 66.93 & 0.047 & \bfseries 79.05 & 0.057 & \bfseries 82.20 & 0.035\\
			\cline{2-8}
			& PDCABB &\bfseries 67.10 & 0.123 & 78.64 & 0.100 & 82.00 & 0.588\\
			\hline
			\multirow{2}*{Clarke+Google+H} & Gurobi & 82.18 & 0.021 &  82.60 & 0.028 & \bfseries 82.63 & 0.023\\
			\cline{2-8}
			& PDCABB &  \bfseries 82.19 & 0.032 & \bfseries 82.63 & 0.037 &   82.60 & 0.041\\
			\hline
	\end{tabular}} 
\end{table}

\begin{figure}[h!]
	\centering
	\includegraphics[width=0.8\textwidth]{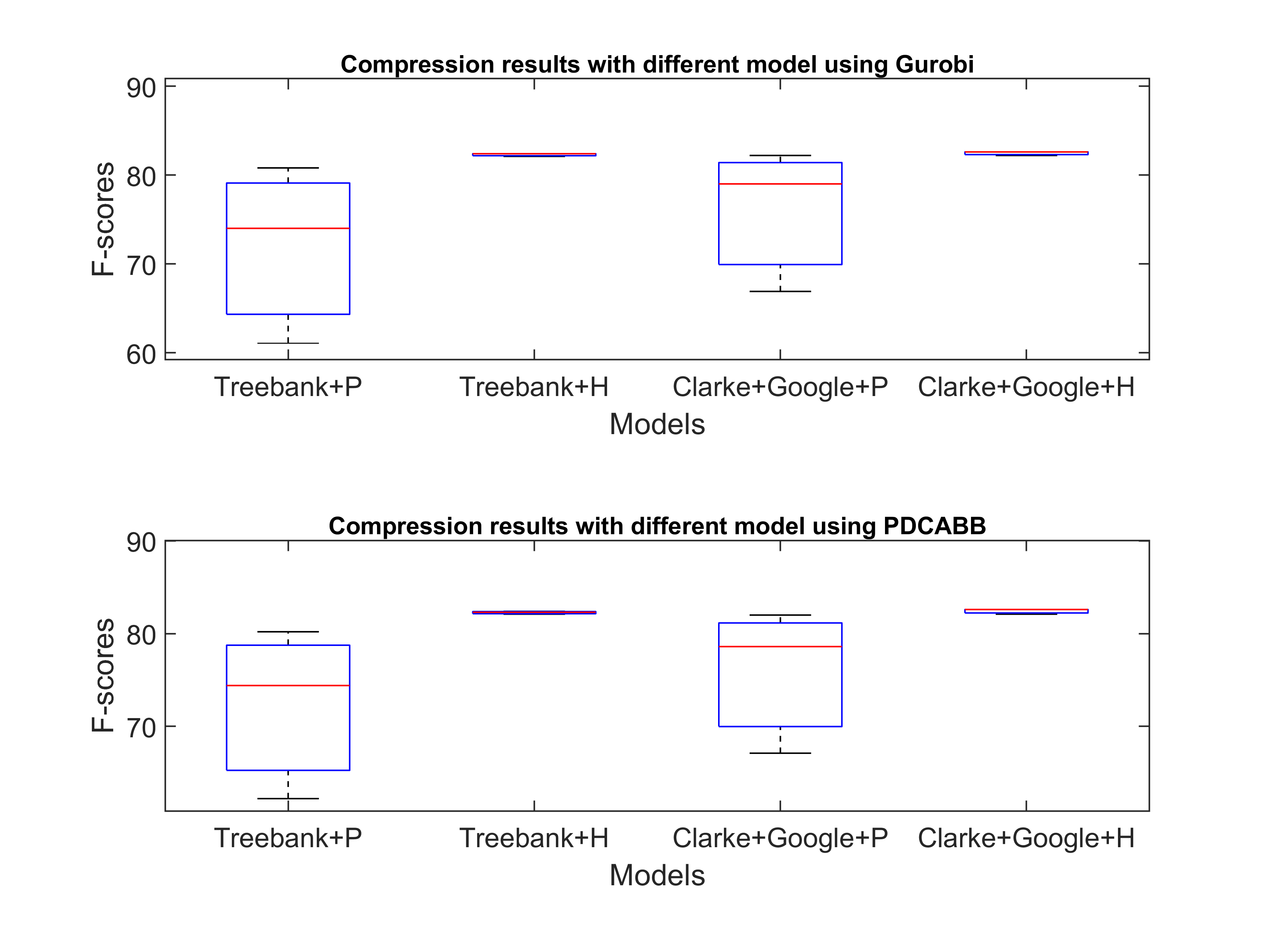}
	\centering  
	\caption{Box-plots for different models v.s. F-scores using Gurobi and PDCABB} \label{fig:boxplot1} 
\end{figure}

The box-plots given in Fig \ref{fig:boxplot1} demonstrates the variations of F-scores for different models with different corpora. We observed that our hybrid model (Treebank+H and Clarke+Google+H) provided better F-scores in average and is more stable in variation, while the quality of the compressions given by the probabilistic model is worse and varies a lot. Moreover, the choice of corpora affect indeed the compression quality since the trigram probability depends on corpora. Therefore, in order to provide more reliable compressions, we have to choose the most related corpora to compute the trigram probabilities.

\begin{figure}[h!]
	\centering
	\begin{minipage}[c]{0.5\textwidth}
		\centering
		\includegraphics[width=\linewidth]{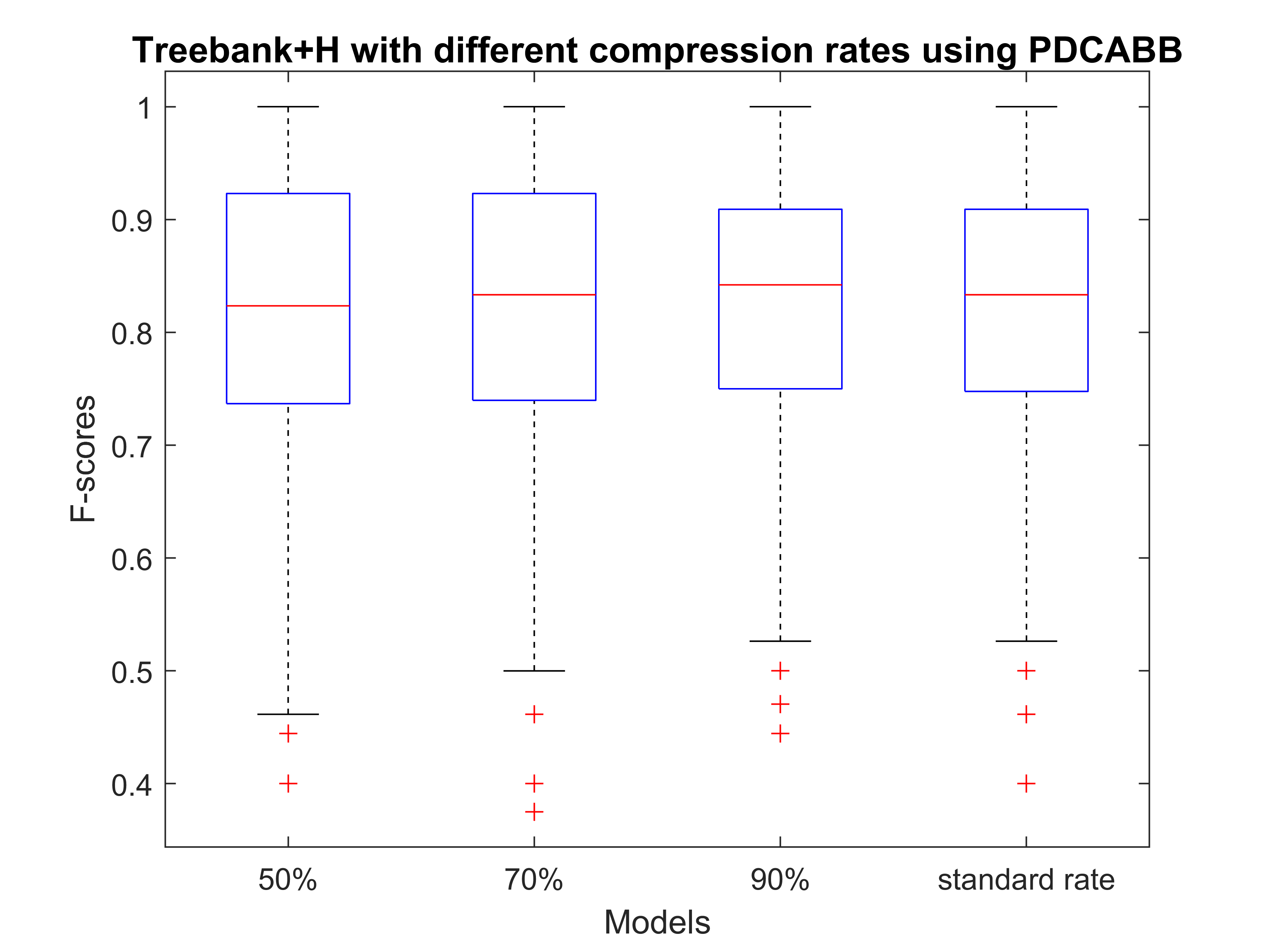}
	\end{minipage}%
	\begin{minipage}[c]{0.5\textwidth}
		\centering
		\includegraphics[width=\linewidth]{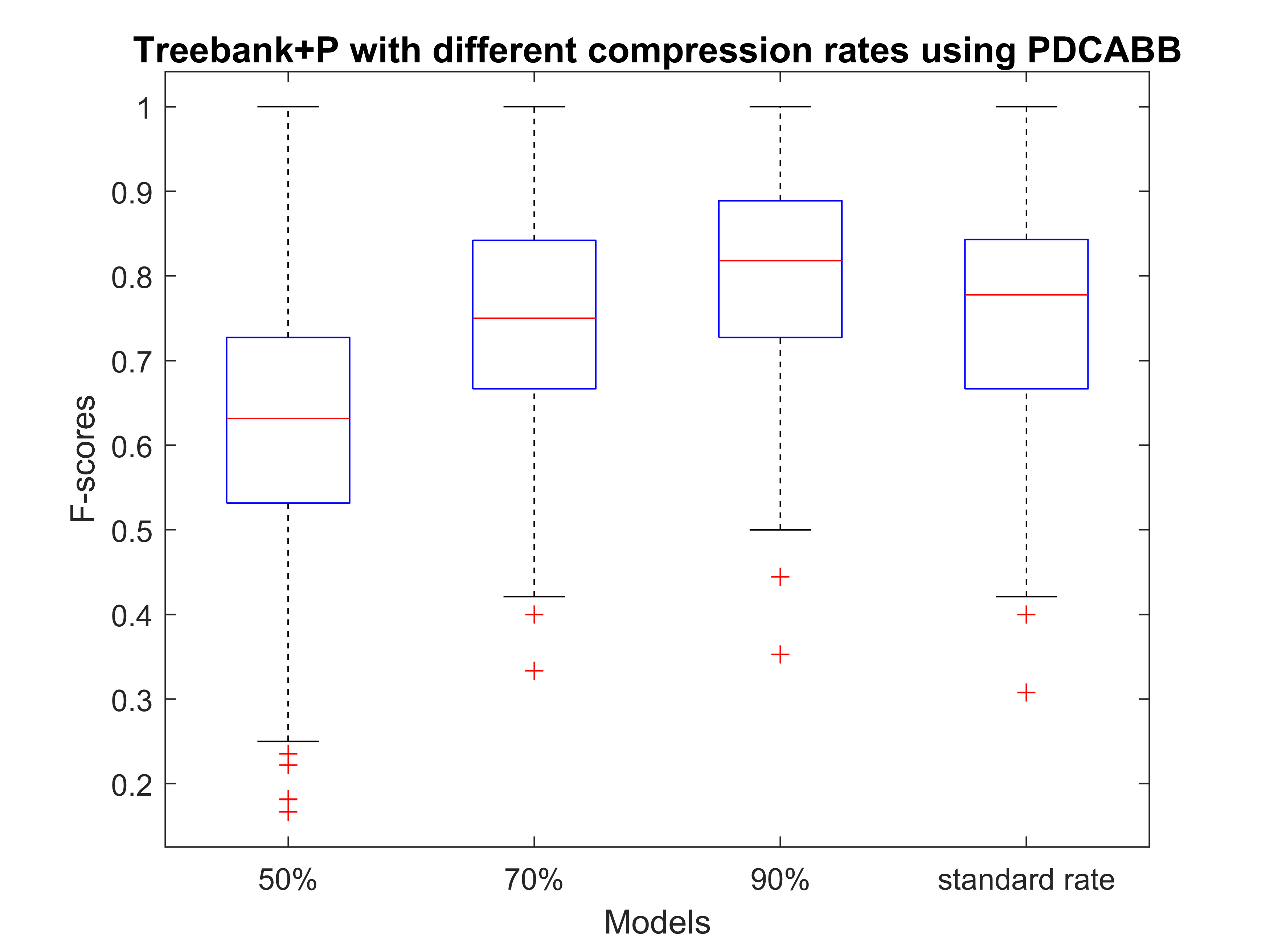}
	\end{minipage}\\
	\begin{minipage}[c]{0.5\textwidth}
		\centering
		\includegraphics[width=\linewidth]{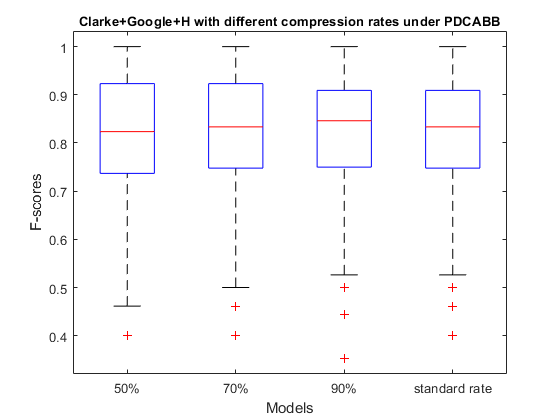}
	\end{minipage}%
	\begin{minipage}[c]{0.5\textwidth}
		\centering
		\includegraphics[width=\linewidth]{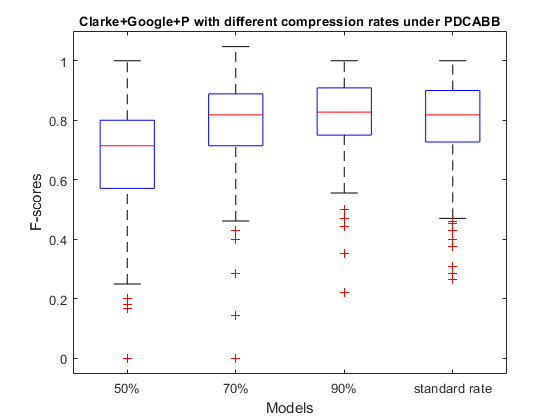}
	\end{minipage}\\
	\caption{Box plots of compression rates v.s. F-scores for different models using PDCABB}
	\label{fig:box}
\end{figure}
The box plots of compression rates v.s. F-scores for different models using PDCABB is illustrated in Fig \ref{fig:box}. We present here the numerical results of PDCABB only since Gurobi gets very similar results without visual differences in the figure. The values of F-scores are also computed with particular compression rate obtained from the standard compression, namely \emph{standard compression rate}, whose results are very similar to the results with $70 \%$ compression rate. It seems that the worst F-scores are found around $50\%$ compression rate, and the best compressions appear with $90\%$ compression rate for all models. Moreover, our hybrid model always get compressions with higher F-scores than the probabilistic model. 

As a conclusion, our hybrid model outperforms the probabilistic model in the compression quality no matter what corpus used. PDCABB algorithm can solve our hybrid model very efficiently to provide stable and high F-score compression results. It seems to be a promising approach for sentence compression.

\section{Conclusion and Perspectives}\label{sec:conclusion_perspectives}

We have proposed a hybrid sentence compression model ILP-PT based on the probabilistic model and the parse tree model to guarantee the syntax correctness of the compressed sentence and save the main meaning. We use a DC programming approach PDCABB for solving our sentence compression model. Experimental results show that our new model and the solution algorithm can produce high quality compressed results within a short compression time.
	
Concerning on future work, we are very interested in using recurrent neural network for the sentence compression task and particularly for sentence structure classification and grammar generation. On the other hand, recently, we have developed in \cite{paper_niu_2019} a parallel cutting plane algorithm based on the new DC cut and the classical global cuts (such as lift-and-project cut and Gomory's cut) combining with DCA for globally solving problem \eqref{P}. So it is interesting to test its performance in sentence compression and to compare with PDCABB. Researches in these directions will be reported subsequently.

\begin{acknowledgements}
	The authors are supported by the National Natural Science Foundation of China (Grant 11601327) and the National ``$985$" Key Program of China (Grant WF220426001).
\end{acknowledgements}

%
%

\bibliographystyle{spmpsci}      
\bibliography{refs}   


\section*{Appendix}
\begin{appendix}
	\section{List of Labels for POS Tagging and Parsing}\label{appendix:A}
	Some labels used for sentence parsing and POS tagging are listed in Table \ref{tab:labels}. Some notations refer to POS tags of the Penn Treebank Tagset \cite{NLTK}, others are introduced by ourselves. For example, notation ``ADJ'' refers to adjective tag cases ``JJ'', ``JJR'' and ``JJS'' in the Penn Treebank Tagset, but ``ADJP'' is a notation that we defined to denote `adjective phrase'.
	
	\begin{longtable}[htbp!]{lp{3cm}lp{3.6cm}}
		\caption{Labels for POS tagging and parsing}\label{tab:labels}\\
		\hline
		Tag  & Penn Treebank Tagset & Meaning  &  Examples \\
		\hline
		ADJ  & JJ, JJR, JJS &adjective  &  new, good, high \\
		ADJP  & -- &adjective phrase  &  very cute, extremely tasty \\
		ADV  & RB, RBR, RBS &adverb  &  really, already, still \\
		ADVC  & -- &  adverbial clause  & While he was sleeping, (his wife was cooing.)\\
		ADVP  & -- &  adverb phrase  & clear enough\\
		ATTC  & -- & attributive clause  & \\
		CC  & CC & conjunction &  and, or, but \\
		CD  & CD &number,cardinal  &  mid-1890, nine-thirty\\
		CONJP  & -- &  conjunction phrase  & and, as well as\\
		DT  & DT & determiner  &  the, some, this \\
		EX  & EX & existential there  &  there\\
		IN  & IN & preposition  &  in, of, with \\
		N  & NN, NNP, NNPS, NNS &noun  &  year, home, time \\
		NP  & -- &noun phrase  &  milk tea, sentence compression \\
		OC  & -- &  object clause  & (He promises) that he will come back.\\
		P  & PRP, PRP\$, WP, WP\$ &pronoun  &  he, she, you \\
		PP  & -- &preposition phrase  &  in the park, with a book \\
		QP  & -- &quantifier phrase  &  more than one \\
		S  & -- &sentence  & \\
		SBAR  & -- &  clause  & \\
		SC  & -- &  subject clause  & What I said (is right.)\\
		SYM  & `.', `,', `!', `?', `;', `:' &symbol  &  `.', `,', `!', `?', `;', `:' \\
		TO  & TO &the word to  &  to \\
		TOP  & -- &to do phrase  &  to eat, to have fun \\
		V  & MD, VB, VBD, VBG, VBN, VBP, VBZ &verb  &   is, has, get \\
		VP  & -- &verb phrase  &  have lunch, eat cake \\
		WDT  & WDT &WH determiner  &  which, what, whichever \\
		WP  & WP &  WH-pronoun  & that, whatever, which, who\\
		WRB  & WRB &  WH-adverb  & how, however, where, why\\
		\hline
	\end{longtable}
	
	\section{List of CFG Grammar and Compression Rules for Statements}\label{appendix:B}
	\begin{longtable}[htbp!]{llp{5cm}}
		\caption{CFG grammar and compression rules for statements}\label{tab:compressionrules}\\
		\hline
		Grammars & Compression Rules & Examples\\
		\hline
		S$\to$(NP VP SYM) & (1 1 1)& She bought a new book.\\
		S$\to$(NP VP) & (1 1)& (That is great if) you come here(.)\\
		\hline
		NP$\to$(N) & (1)& tea\\
		NP$\to$(N NP) & (1 1)& milk tea\\
		NP$\to$(N PP) & (1 2)& tea with sugar\\
		NP$\to$(N ATTC) & (1 2)& (This is the) tea which he offered. \\
		NP$\to$(N SBAR) & (1 2)& a situation in which...\\
		NP$\to$(SC) & (1)& Whether we can win (is still unknown.)\\
		NP$\to$(N CC NP) & (1 1 1)& milk and tea\\
		NP$\to$(N ADVP) & (1 2)& (The) woman there (is your mother.)\\
		NP$\to$(DT) & (1)& this\\
		NP$\to$(DT NP) & (2 1)& this book\\
		NP$\to$(DT ADJP) & (2 1)& the rich\\
		NP$\to$(EX) & (1)& there\\
		NP$\to$(ADJP NP) & (2 1)& new book\\
		NP$\to$(CC NP) & (1 1)& and the book\\
		NP$\to$(CD NP) & (2 1)& two books\\
		NP$\to$(QP NP) & (2 1)& more than one book\\
		NP$\to$(P) & (1)& he\\
		NP$\to$(P NP) & (1 1)& my book\\
		NP$\to$(N TOP) & (1 1)& book to read\\
		\hline
		VP$\to$(V) & (1) & eat\\
		VP$\to$(V IN OC) & (1 1 1) & (Our success) depends on how well we can cooperate with others.\\
		VP$\to$(V IN NP) & (1 1 1) & depend on you\\
		VP$\to$(V NP) & (1 1) & have dinner\\
		VP$\to$(V VP) & (1 1) & (he) is writing\\
		VP$\to$(V OC) & (1 1) & (I) heard that he joined the army.\\
		VP$\to$(V P OC) & (1 1 1) & (She) told me that she was beautiful.\\
		VP$\to$(V NP VP) & (1 1 1) & make the baby eat\\
		VP$\to$(V NP PP) & (1 1 2) & have dinner in the restaurant\\
		VP$\to$(ADVP VP) & (1 1) & happily eat\\
		VP$\to$(V ADVP) & (1 1) & eat happily\\
		VP$\to$(V ADVP PP) & (1 2 2) & eat happily in the restaurant\\
		VP$\to$(V ADVP NP) & (1 2 1) & play happily the piano\\
		VP$\to$(V PP) & (1 2) & eat in the restaurant\\
		VP$\to$(V PP PP) & (1 2 2) & eat in the restaurant at noon\\
		VP$\to$(V TOP) & (1 1) & stop to eat\\
		VP$\to$(V ADJP) & (1 1) & (the book) is nice\\
		VP$\to$(V ADVP ADVC) & (1 1 2) & (I) come back late because I was on duty.\\
		VP$\to$(V ADJP ADVC) & (1 1 2) & (He) is smart because he read a lot.\\
		\hline
		ADJP$\to$(ADJ) & (2) & beautiful\\
		ADJP$\to$(ADV ADJ) & (2 1) & extremely fantastic\\
		ADJP$\to$(ADJ OC) & (1 2) & (I am) afraid that I've made a mistake.\\
		\hline
		ADVP$\to$(ADV) & (2) & happily\\
		ADVP$\to$(ADJ ADV) & (2 2) & clear enough\\
		ADVP$\to$(ADV ADV) & (2 1) & (work) extremely hard\\
		\hline
		CONJP$\to$(IN ADV IN) & (2 2 2) & as well as\\
		CONJP$\to$(CC) & (1) & and, or\\
		\hline
		PP$\to$(IN NP) & (1 1) & in the park\\
		PP$\to$(IN ADJ) & (1 1) & at last\\
		PP$\to$(IN NP IN) & (1 1 1) & (he worked) in this company before\\
		PP$\to$(IN CD NP) & (1 1 1) & in five minutes\\
		\hline
		ATTC$\to$(P VP) & (2 2) & (This is the present) he gave me\\
		ATTC$\to$(WDT S) & (2 2) & (He is the one) who spoke at the meeting.\\
		ATTC$\to$(WRB S) & (2 2) & (This is the place) where we met.\\
		\hline
		ADVC$\to$(IN S) & (1 1) & (I came late) because I was on duty.\\
		\hline
		OC$\to$(IN S) & (1 1) & (I heard) that he joined the army.\\
		OC$\to$(WP VP) & (1 1) & (She didn't know) what had happened.\\
		OC$\to$(WRB TOP) & (1 1) & (I know) where to go.\\
		OC$\to$(WRB ADV S) & (1 1 1) & (Our success depends on) how well we can cooperate with others.\\
		OC$\to$(WP S) & (1 1) & (I didn't know) where they were born.\\
		\hline
		SC$\to$(IN S) & (1 1) & Whether we can win (is still unknown.)\\
		SC$\to$(WP S) & (1 1) & What I want (are two books.)\\
		SC$\to$(WDT VP) & (1 1) & Whichever comes first (will receive a prize.)\\
		SC$\to$(WDT PP VP) & (1 2 1) & Whichever of you comes in first (will receive a prize.)\\
		SC$\to$(WRB S) & (1 1) & How it was done (was a mystery.)\\
		\hline
		TOP$\to$(TO VP) & (1 1) & to have dinner\\
		\hline
		QP$\to$(ADJ IN CD) & (1 1 1) & more than one\\
		QP$\to$(IN CD N) & (1 1 1) & about 50 dollars\\
		QP$\to$(CD IN N) & (1 1 1) & one in fifth\\
		QP$\to$(DT N IN) & (1 1 1) & a quarter of, a majority of, a number of\\
		QP$\to$(DT ADJ N IN) & (1 1 1) & a large number of\\
		QP$\to$(ADV DT ADJ) & (1 1 1) & quite a few\\
		QP$\to$(N IN) & (1 1 1) & plenty of\\
		QP$\to$(CD N IN) & (1 1 1) & twenty percent of\\
		\hline
		SBAR$\to$(WDT S) & (1 1) & (This is the book) which he gave me.\\
		\hline
	\end{longtable}
\end{appendix}

\end{document}